\documentclass[conference]{IEEEtran}
\IEEEoverridecommandlockouts

\usepackage{url}
\usepackage{cite}
\usepackage{amsmath,amssymb,amsfonts}
\usepackage{algorithmic}
\usepackage{graphicx}
\usepackage{textcomp}
\usepackage{xcolor}

\def\BibTeX{{\rm B\kern-.05em{\sc i\kern-.025em b}\kern-.08em
    T\kern-.1667em\lower.7ex\hbox{E}\kern-.125emX}}

\begin{document}

\title{Cross-Dataset Generalization in Breast MRI Tumor Classification via Class-Wise Dataset Mixing}

\author{\IEEEauthorblockN{Mohammad Ali Dadrast}
\IEEEauthorblockA{\textit{Data Science Program} \\
\textit{Memorial University of Newfoundland}\\
St. John's, Canada \\
madadrast@mun.ca}
\and
\IEEEauthorblockN{Hamid Usefi}
\IEEEauthorblockA{\textit{Department of Mathematics and Statistics} \\
\textit{Memorial University of Newfoundland}\\
St. John's, Canada \\
usefi@mun.ca}
}

\maketitle

\begin{abstract}
Breast MRI is highly sensitive for detecting breast tumors, but exams contain many slices and require substantial reading time. Deep learning models often perform well on internal splits but can fail across institutions because of domain shift and dataset-origin bias. We study this failure mode for binary breast MRI tumor classification. EfficientNet-B3 and WaveViT-Small are trained using Duke Breast Cancer MRI and fastMRI, and evaluated only on the independent multi-center MAMA-MIA cohort. In a deliberately confounded setup, where label is perfectly correlated with dataset origin, external accuracy is near chance (0.5048--0.5265), despite very high recall. We then construct a mixed training set in which each class contains samples from both Duke and fastMRI, while preserving patient-level splitting, augmentation, and leakage controls. On MAMA-MIA, dataset mixing improves accuracy/F1 to 0.8463/0.8625 for WaveViT-Small and 0.8884/0.8994 for EfficientNet-B3. These results show that controlling dataset-origin bias is important for reliable breast MRI classification.
\end{abstract}

\begin{IEEEkeywords}
Breast cancer, MRI, cross-dataset validation, domain generalization, dataset bias, classification.
\end{IEEEkeywords}

\section{Introduction}
Breast cancer is one of the most common cancers among women worldwide, with more than 2.3 million new cases annually and a large clinical and economic burden \cite{WHO2026}. Earlier diagnosis is associated with better prognosis and lower treatment cost \cite{WHO2026,Sun2018}. Breast MRI is important for high-risk screening, staging, and treatment planning because of its high sensitivity, but one examination can contain hundreds or thousands of images across sequences and time points. This makes interpretation time-consuming and motivates decision-support tools that can assist radiologists without replacing clinical judgement.

For such tools to be useful, strong internal accuracy is not enough. A classifier should retain its behavior when images are acquired at another site or processed through a different pipeline. This is difficult in MRI because image appearance depends on scanner vendor, field strength, coil configuration, pulse sequence, contrast protocol, reconstruction, and post-processing. A model can therefore learn a local imaging style instead of tumor-related features. This type of shortcut learning has been reported in cross-institutional imaging studies, and external validation reviews show that performance often drops on independent cohorts \cite{Zech2018,Alice2022,Kelly2019}. Recent reporting standards also emphasize transparent evaluation and external validation \cite{Collins2024,Sounderajah2025}.

Breast MRI is especially vulnerable because public datasets differ not only in patient population but also in acquisition protocol, annotation style, and release format. Duke Breast Cancer MRI provides DICOM images with lesion bounding boxes \cite{TCIADukeBreastMRI}, fastMRI provides HDF5 volumes and case-level labels for reconstruction research \cite{Zbontar2018,Solomon2025}, and MAMA-MIA provides a multi-center DCE-MRI benchmark with lesion segmentations \cite{Garrucho2025}. These differences make the datasets useful for testing generalization, but they also create a risk: if one class comes mainly from one dataset and the other class from another dataset, the model may learn dataset origin as a proxy for the label.

In this work, we study cross-dataset generalization for breast MRI tumor classification. EfficientNet-B3 \cite{Tan2019} and WaveViT-Small \cite{Yao2022} are trained using Duke and fastMRI, while MAMA-MIA is reserved for strict external validation. We first create a deliberately confounded training set to expose dataset-origin shortcut learning. We then use class-wise dataset mixing, so that both labels contain samples from both Duke and fastMRI. The architecture and training pipeline are kept fixed, which makes the effect of the training-data composition easier to interpret.

\section{Related Work}
Deep learning is widely used in breast image analysis, and both convolutional and transformer-based backbones have reported strong results when training and testing are performed within the same distribution \cite{Tan2019,Yao2022}. In breast MRI, the combination of high soft-tissue contrast and multiphase acquisition provides useful diagnostic information, but it also increases sensitivity to acquisition and processing choices that are not disease-specific.

The key issue is the gap between internal accuracy and external reliability. Internal train--test splits are useful for model development, but they cannot rule out source-specific learning when all images come from one environment. A stricter test is to hold out a cohort collected under different institutional and technical conditions. Cross-institutional studies have shown that imaging models can learn hospital or system signatures instead of disease mechanisms \cite{Zech2018}. Reviews of external validation studies also report that independent testing remains less common than internal testing and that performance often drops when external cohorts are used \cite{Alice2022}.

Breast MRI makes this problem more pronounced because public datasets were created for different purposes. Duke is a curated TCIA collection with lesion localization and has supported prior radiogenomics work \cite{Saha2018,TCIADukeBreastMRI}. fastMRI was designed mainly for accelerated reconstruction research \cite{Zbontar2018,Solomon2025}. MAMA-MIA was introduced as a large multi-center DCE-MRI benchmark with expert-corrected segmentations \cite{Garrucho2025}. Our work uses these differences deliberately: rather than treating dataset heterogeneity only as a nuisance, we use it to test whether models learn transferable tumor cues or dataset-origin shortcuts.

\section{Materials and Methods}

\subsection{Datasets}
We use three public breast MRI datasets collected by different institutions and released in different formats. Duke contains dynamic contrast-enhanced MRI studies from 922 patients with biopsy-confirmed invasive breast cancer, collected at Duke University between 2000 and 2014 \cite{TCIADukeBreastMRI,Clark2013}. The scans were acquired on 1.5T and 3T scanners and released in DICOM format. Lesions were annotated by trained readers using bounding boxes, making Duke the main source of well-localized positive slices.

The fastMRI project provides raw and reconstructed MRI data collected across heterogeneous hardware and acquisition settings \cite{Zbontar2018,Solomon2025}. We used the approved breast MRI subset, distributed as multi-coil HDF5 volumes with reconstructed images. Since fastMRI was designed mainly for reconstruction, detailed lesion annotations are not central to the release. In the curated subset used here, there were 300 patients: 51 negative, 159 benign, and 90 malignant. We used only negative and malignant cases to keep the task binary.

MAMA-MIA contains 1,506 pre-treatment DCE-MRI studies from multiple clinical centers in the I-SPY2 trial \cite{Garrucho2025}. The scans are released in compressed NIfTI format and include both unilateral and bilateral studies, with fat-suppressed and non-fat-suppressed protocols. Tumor and non-mass-enhancement segmentations were generated automatically and quality-controlled by breast imaging specialists. Because of its multi-center design and scanner diversity, MAMA-MIA was used only for external validation.

\subsection{Data Preparation}
The datasets differ in file format, annotation type, and labeling granularity. Each dataset was therefore curated separately and then converted to a common slice-based representation. After slice selection, images were resized to $256\times256$, converted to three-channel PNG format, and labeled as non-tumor (0) or tumor-containing (1). Standardized filenames preserved patient identifiers and slice indices. Train--test splitting for Duke and fastMRI was performed at the patient level using an 80/20 ratio, so slices from the same patient never appeared in both splits.

\subsubsection{Preparing Duke MRI Dataset}
Duke volumes were read from DICOM files and linked to the released bounding boxes. Slices intersecting annotated tumor regions were extracted as positive samples, while healthy slices were sampled away from tumor regions. To reduce noise and possible mislabeling, the first and last ten healthy slices were removed, together with borderline slices near tumor boundaries. Healthy slices were downsampled for class balance. Random rotations and vertical flips were used during training, and the final images were loaded through PyTorch DataLoaders with batch size 64 \cite{TCIADukeBreastMRI}. A sample Duke slice is shown in Fig.~\ref{fig:mri_samples}(a).

\subsubsection{Preparing MAMA-MIA (ISPY2) MRI Dataset}
For MAMA-MIA, MRI scans and segmentation masks were aligned, and axial slices containing segmented tumor regions were extracted and converted to $256\times256$ three-channel PNG images. We retained cases with adequate contrast quality, no major artifacts, bilateral scans, and no breast implants. Healthy and tumor-containing slices were selected from each patient to reduce label uncertainty. Labels were assigned using the expert-corrected segmentations, and MAMA-MIA was not used for training, tuning, or internal testing. A sample MAMA-MIA slice is shown in Fig.~\ref{fig:mri_samples}(b).

\subsubsection{Preparing fastMRI Breast Dataset}
\begin{figure*}[t]
	\centering
	\begin{tabular}{@{}ccc@{}}
		\includegraphics[width=.31\textwidth]{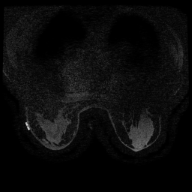} &
		\includegraphics[width=.31\textwidth]{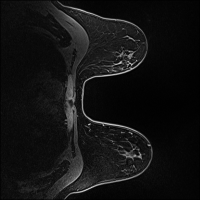} &
		\includegraphics[width=.31\textwidth]{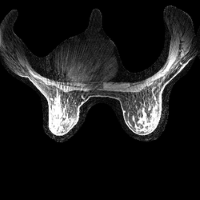} \\
		\scriptsize (a) Duke & \scriptsize (b) MAMA-MIA & \scriptsize (c) fastMRI
	\end{tabular}
	\caption{Sample slices from the three breast MRI datasets.}
	\label{fig:mri_samples}
\end{figure*}

The fastMRI breast data were processed from multi-coil HDF5 volumes. After access approval, volumes were extracted at the patient level according to standard fastMRI reconstruction protocols \cite{Zbontar2018,Solomon2025}. Each volume was normalized using percentile-based intensity scaling. Axial slices were converted to PNG format, resized to $256\times256$, and saved with patient identifiers and slice indices. Negative cases contributed central breast slices (indices 20--170), while malignant cases contributed slices around the estimated tumor center using mapping files. Benign cases were excluded. A sample fastMRI slice is shown in Fig.~\ref{fig:mri_samples}(c).

The same naming and splitting rules were used in all experiments. In particular, source files were never split at the slice level before patient grouping. This was important because adjacent MRI slices are highly correlated and can otherwise produce overly optimistic estimates. Augmentation was applied only after the training split was defined, and the external MAMA-MIA cohort was kept separate from all choices related to model training, tuning, and class balancing.

\subsection{Model Architectures}
We evaluated EfficientNet-B3 and WaveViT-Small. EfficientNet-B3 is a convolutional architecture based on compound scaling, mobile inverted bottleneck blocks, and squeeze-and-excitation modules \cite{Tan2019}. WaveViT-Small combines discrete wavelet transforms with self-attention to capture local detail and broader spatial context \cite{Yao2022}. Both models were configured for binary classification with input size $256\times256$. Training used AdamW, learning rate $10^{-4}$, weighted cross-entropy loss, batch size 64, and a ReduceLROnPlateau scheduler.

\section{Experiments}
The experiments ask whether models trained on multiple breast MRI sources learn tumor-related features or exploit source-specific cues. We first create a deliberately confounded dataset to quantify dataset--label correlation. We then use class-wise dataset mixing to break this correlation while keeping the models, loss, optimizer, augmentation, and evaluation pipeline fixed. MAMA-MIA is used only for external validation.

\subsection{Baseline: Confounded Multi-Source Training}
We created a confounded dataset by combining malignant slices from Duke with negative slices from fastMRI. Duke malignant slices were assigned label 1 and fastMRI negative slices were assigned label 0. Images were renamed using a common convention based on patient identifier and slice index. This setup creates a worst-case confound: class label is perfectly aligned with dataset origin. The aim is not to build a deployable classifier, but to test whether a standard pipeline can be misled into learning source identity rather than pathology.

\begin{figure}[htbp]
	\centerline{\includegraphics[width=.96\linewidth]{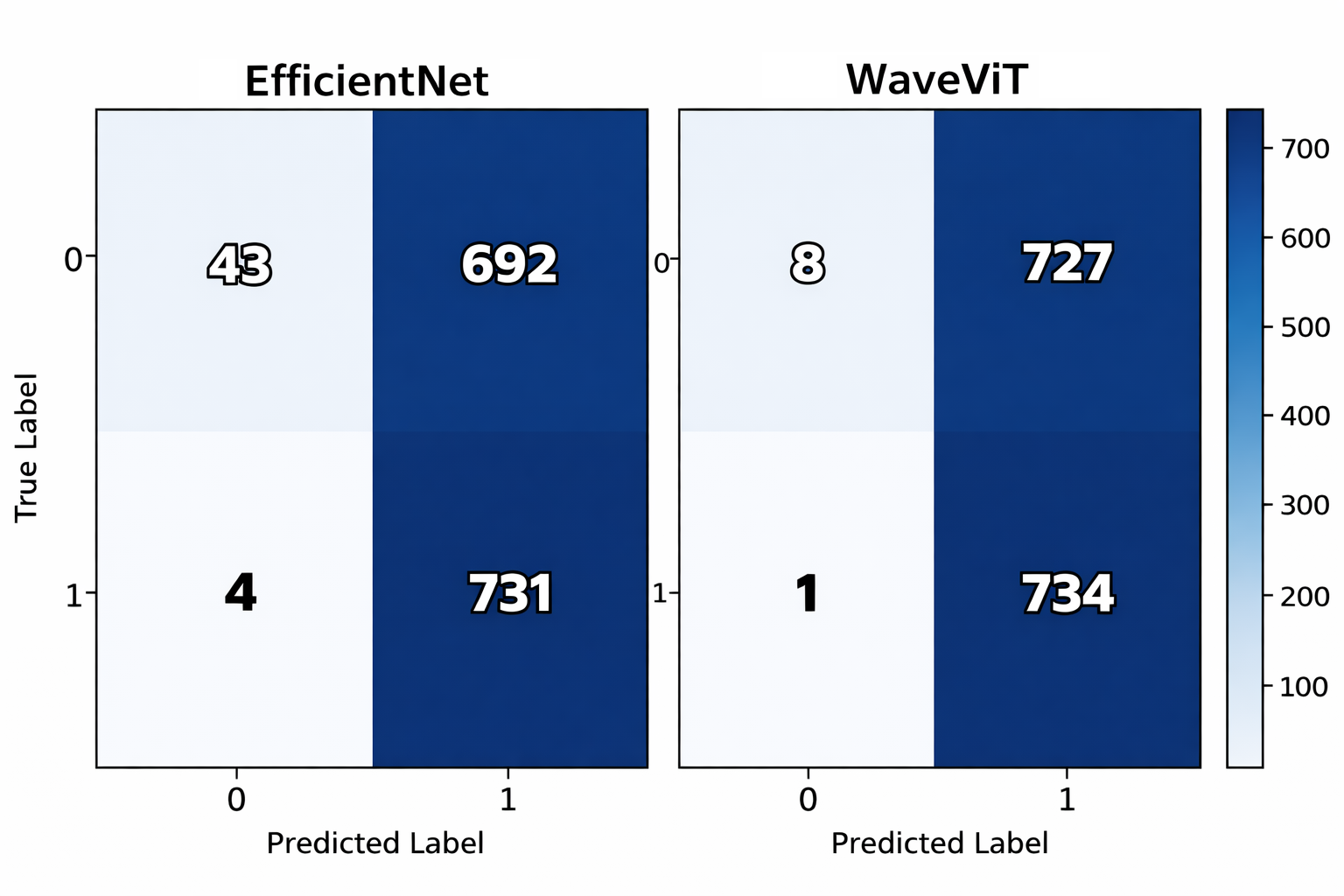}}
	\caption{Confusion matrices for the confounded New Dataset.}
	\label{image1}
\end{figure}

WaveViT-Small and EfficientNet-B3 were trained on this New Dataset using patient-level splitting, augmentation, and leakage controls. After training, both models were evaluated on MAMA-MIA as a fully independent external validation set. Table~\ref{external_results} reports the results, and Fig.~\ref{image1} shows the confusion matrices.

\begin{table}[htbp]
	\caption{External performance on MAMA-MIA.}
	\begin{center}
	\scriptsize
	\resizebox{\columnwidth}{!}{%
		\begin{tabular}{|c|c|c|c|c|c|}
			\hline
			\textbf{Training set} & \textbf{Model} & \textbf{Accuracy} & \textbf{Precision} & \textbf{Recall} & \textbf{F1}  \\
			\hline
			Confounded & WaveViT-Small   & 0.5048 & 0.5024 & 0.9986 & 0.6685 \\
			\hline
			Confounded & EfficientNet-B3 & 0.5265 & 0.5137 & 0.9946 & 0.6775 \\
			\hline
			Mixed & WaveViT-Small      & 0.8463 & 0.7800 & 0.9646 & 0.8625 \\
			\hline
			Mixed & EfficientNet-B3    & 0.8884 & 0.8190 & 0.9973 & 0.8994 \\
			\hline
		\end{tabular}}
		\label{external_results}
	\end{center}
\end{table}

Both models fail externally: recall is near one, but accuracy is close to chance. This means that the networks mostly predict the positive class on MAMA-MIA. The most plausible explanation is the confounding in the New Dataset: all positives come from Duke and all negatives come from fastMRI, so acquisition, preprocessing, or reconstruction signatures can serve as shortcuts for the label. These shortcuts do not transfer to MAMA-MIA, where dataset origin is unrelated to the class label.

\subsection{Proposed Approach: Class-Wise Dataset Mixing}
To reduce dataset-origin bias, we construct a \emph{Mixed Dataset} in which both classes contain samples from both Duke and fastMRI. For label 1, we include all Duke positive slices and add fastMRI positive slices equal to 40\% of the Duke positive count, or all available fastMRI positives if fewer. For label 0, we include all fastMRI negative slices and add Duke negative slices equal to 40\% of the fastMRI negative count, or all available Duke negatives if fewer. The 40\% value adds cross-source variation without letting the smaller source dominate the class composition.

Patient-level splitting is then performed with an 80/20 train--test ratio, followed by the same training pipeline used in the confounded experiment. Table~\ref{mix_table} gives the resulting slice counts.

\begin{table}[htbp]
	\caption{Mixed Dataset after patient-level splitting.}
	\begin{center}
	\scriptsize
		\begin{tabular}{|c|c|c|c|}
			\hline
			\textbf{Split / Label} & \textbf{Duke} & \textbf{fastMRI} & \textbf{Total} \\
			\hline
			Test, Label 1  & 4355 & 918  & 5273 \\
			\hline
			Test, Label 0  & 664  & 1661 & 2325 \\
			\hline
			Train, Label 1 & 19071 & 3672 & 22743 \\
			\hline
			Train, Label 0 & 2416  & 6040 & 8456 \\
			\hline
		\end{tabular}
		\label{mix_table}
	\end{center}
\end{table}

\begin{figure}[htbp]
	\centerline{\includegraphics[width=.96\linewidth]{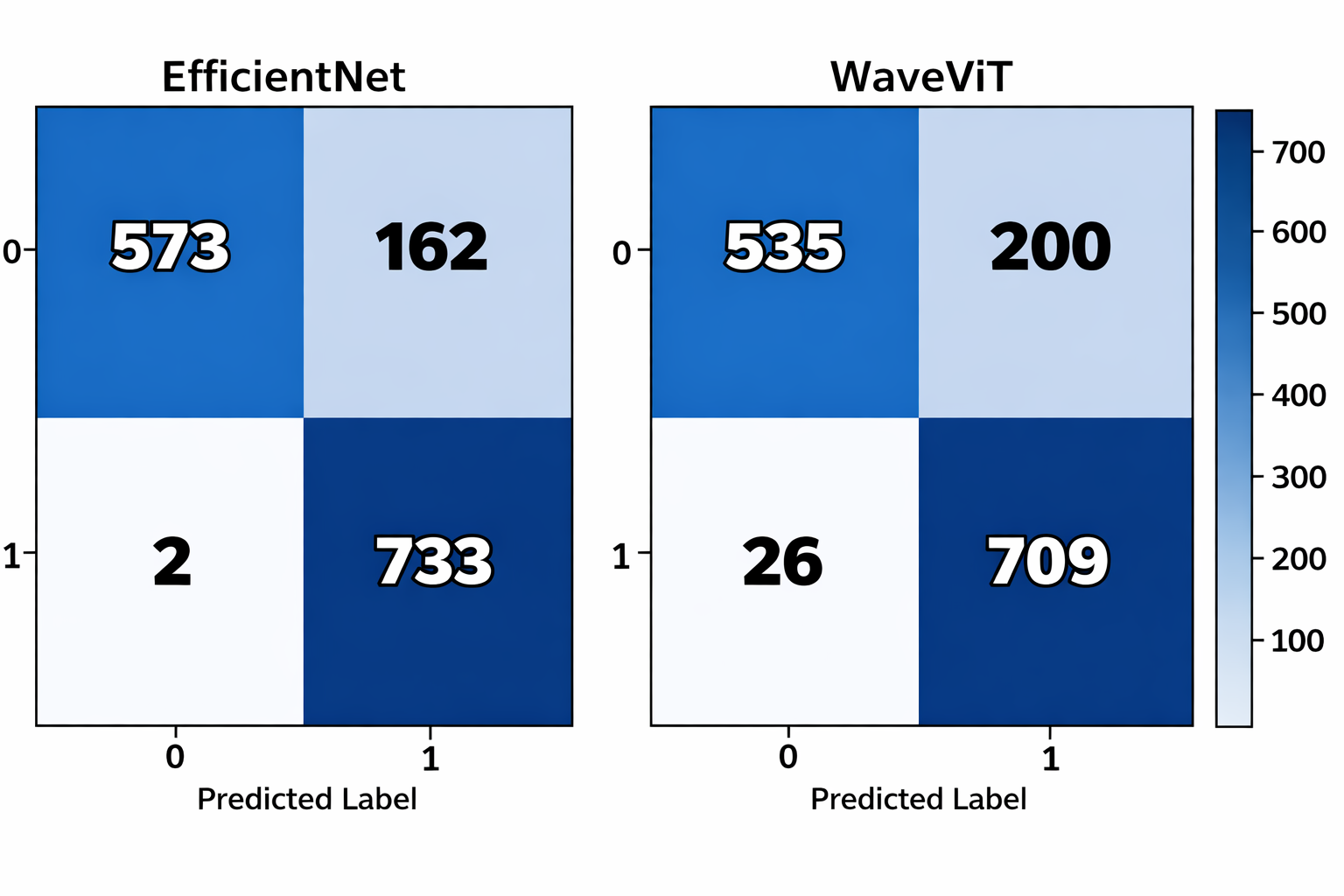}}
	\caption{Confusion matrices for the Mixed Dataset.}
	\label{image2}
\end{figure}

Both models were retrained on the Mixed Dataset and evaluated on MAMA-MIA using the same external protocol. The results are included in Table~\ref{external_results}, and the confusion matrices are shown in Fig.~\ref{image2}.

Class-wise mixing substantially improves external generalization. WaveViT-Small improves from 0.5048 to 0.8463 in accuracy and from 0.6685 to 0.8625 in F1-score. EfficientNet-B3 improves from 0.5265 to 0.8884 in accuracy and from 0.6775 to 0.8994 in F1-score, while maintaining very high recall. Precision also improves, indicating that the models no longer classify nearly all external slices as positive. These gains are obtained without changing the architecture, loss, or optimizer, so they support the main claim that source-aware data composition can reduce dataset-origin shortcut learning.

\section{Conclusion}
This work studied breast MRI tumor classification under cross-dataset shift using WaveViT-Small and EfficientNet-B3. The confounded experiment showed a clear failure mode: when all positives came from Duke and all negatives came from fastMRI, both models achieved near-chance external accuracy on MAMA-MIA while predicting most slices as positive. This pattern is consistent with dataset-origin shortcut learning.

Class-wise mixing reduced this bias by ensuring that both labels contained samples from both Duke and fastMRI. With the same architectures and training pipeline, external performance improved substantially on MAMA-MIA. EfficientNet-B3 reached 0.8884 accuracy and 0.8994 F1-score, while WaveViT-Small reached 0.8463 accuracy and 0.8625 F1-score. These results suggest that the way public datasets are combined can be as important as the choice of backbone.

The main goal was therefore not to maximize a deployment-ready patient-level score, but to identify whether the slice-level classifier was relying on a transferable signal.
All results are based on per-slice predictions. This is conservative because several correlated slices come from the same patient, and difficult errors often occur near lesion boundaries or low-contrast regions. Patient-level prediction could aggregate evidence across slices by averaging probabilities, majority voting, top-$k$ pooling, or learned aggregation. We leave this step for future work because the focus here was to isolate dataset-origin bias. Overall, the results support source-aware class-wise multi-source training and strict external validation as practical methodological requirements for more reliable clinical breast MRI AI systems.

\end{document}